\documentclass[11pt,letterpaper]{article}

\usepackage[nohyperref]{emnlp2017}\usepackage{xcolor}\definecolor{darkblue}{rgb}{0, 0, 0.5}

\usepackage{times}
\usepackage{latexsym}

\emnlpfinalcopy

\def\emnlppaperid{***}

\usepackage{multirow}
\usepackage{pbox}

\usepackage{url}
\usepackage{amssymb}
\usepackage{amsmath}
\usepackage{amsthm}
\usepackage{graphicx}
\usepackage{subcaption}

\definecolor{myblue}{rgb}{0,0.1,0.6}
\definecolor{mygreen}{rgb}{0,0.3,0.1}
\newcommand{\bocomment}[1]{}
\newcommand{\kkcomment}[1]{}
\newcommand{\ahcomment}[1]{}

\newcommand{\forarxiv}[1]{#1}   

\newenvironment{itemizesquish}{\begin{list}{\labelitemi}{\setlength{\itemsep}{0em}\setlength{\labelwidth}{2em}\setlength{\leftmargin}{\labelwidth}\addtolength{\leftmargin}{\labelsep}}}{\end{list}}
\newcommand{\ignore}[1]{}
\newcommand{\transpose}{^\mathsf{T}}

\usepackage{booktabs}
\usepackage{array}
\newcolumntype{L}{>{$}l<{$}}
\newcolumntype{C}{>{$}c<{$}}
\newcolumntype{R}{>{$}r<{$}}

\title{Identifying civilians killed by police with distantly supervised
entity-event extraction}

\author{Katherine A. Keith, 
Abram Handler, 
Michael Pinkham,
\\
{\bf
Cara Magliozzi,
\ 
Joshua McDuffie, 
\and
Brendan O'Connor} \\
College of Information and Computer Sciences \\
University of Massachusetts Amherst \\
{{\tt kkeith@cs.umass.edu, brenocon@cs.umass.edu}} \\
\url{http://slanglab.cs.umass.edu/}
}

\date{}

\begin{document}

\maketitle

\begin{abstract}
We propose a new, socially-impactful task
for natural language processing: 
from a news corpus, extract names of persons who have been killed by police.
We present a newly collected
police fatality corpus, which we release publicly,
and present a model to solve this problem that uses EM-based
distant supervision
with logistic regression and convolutional neural network classifiers.
Our model outperforms two off-the-shelf event extractor systems,
and it can suggest candidate victim names in some cases
faster
than one of the major manually-collected
police fatality databases.

Appendix, software, and data are available online at: 
 \url{http://slanglab.cs.umass.edu/PoliceKillingsExtraction/}
 
\forarxiv{
    [This paper appears in \emph{Proceedings of EMNLP 2017}. 
    This version includes the appendix.]
}
\end{abstract}

\section{Introduction}
\label{s:intro}

The United States government does not keep systematic records of 
when police kill civilians, despite a clear need for this information to serve 
the public interest and support social scientific analysis.
Federal records rely on incomplete cooperation from local police departments,
and human rights statisticians assess
that they fail to document
thousands of fatalities \citep{Lum2015Homicides}.

News articles have emerged as a valuable alternative data source.
Organizations including
\href{https://www.theguardian.com/us-news/series/counted-us-police-killings}{The Guardian},
\href{https://www.washingtonpost.com/policeshootings/}{The Washington Post},
\href{https://mappingpoliceviolence.org/}{Mapping\ Police\ Violence},
and 
\href{http://www.fatalencounters.org/}{Fatal\ Encounters}
have started to build such databases of U.S. police killings by manually reading
millions of news articles\footnote{Fatal Encounters
director D.~Brian Burghart estimates he and colleagues 
have read 2 million news headlines and ledes
to assemble its fatality records that date back to January, 2000 (pers.~comm.);
we find FE to be the most comprehensive publicly available database.}
and extracting victim names and event details. 
This approach was recently validated by a
Bureau of Justice Statistics study
(\citeauthor{Banks2016ARD}, Dec.~2016)
which
augmented traditional police-maintained records with media reports,
finding twice as many deaths compared to past government analyses.
This suggests textual news data has enormous, real value,
though \emph{manual} news analysis remains extremely laborious.

\begin{table}[t]
\centering
{\footnotesize
  \begin{tabular}{p{5.5cm} p{1.8cm}}
  \toprule
Text & Person killed by police?\\
\midrule
\textbf{Alton Sterling} was killed by  police. & True\\[0.1cm]
Officers shot and killed \textbf{Philando Castile}. & True \\[0.1cm]
Officer \textbf{Andrew Hanson} was shot. & False \\[0.1cm]
Police report \textbf{Megan Short} was fatally shot in apparent murder-suicide. & False\\[0.1cm]
   \toprule
  \end{tabular}
}
\caption{Toy examples (with entities in bold) illustrating the problem of extracting from text names of persons who have been killed by police. 
\label{t:ex}}
\end{table}

We propose to help automate this process by extracting the names of persons killed by police
from event descriptions in news articles (Table~\ref{t:ex}).
This can be formulated as either of
two cross-document entity-event extraction tasks:
\begin{enumerate}
\item Populating an entity-event database: From a corpus of news articles $\mathcal{D}^{(test)}$ over timespan $T$, extract the names of persons killed by police during that same timespan ($\mathcal{E}^{(pred)}$).
\item Updating an entity-event database: In addition to $\mathcal{D}^{(test)}$, assume access to both a
historical database of killings $\mathcal{E}^{(train)}$ and a historical news corpus $\mathcal{D}^{(train)}$ 
for events that occurred before $T$.  This setting often occurs in practice, and is the focus of this paper;
it allows for the use of distantly supervised learning methods.\footnote{\citet{Konovalov2017Events} studies the database update task where edits to Wikipedia infoboxes constitute events.}
\end{enumerate}
\noindent
The task itself has important social value, 
but the NLP research community may be interested
in a scientific justification as well.
We propose that police fatalities are a useful test case for event extraction research.
Fatalities are a well defined type of event with clear semantics for coreference,
avoiding some of the more complex issues in this area \citep{Hovy2013Events}.
The task also builds on a considerable information extraction literature
on knowledge base population 
(e.g.\ \citet{Craven1998WebKB}).
Finally, 
we posit that
the field of natural language processing should, when possible, advance applications of
important public interest.
Previous work established the value of textual news for this problem,
but computational methods could alleviate the scale of manual labor needed to use it.

To introduce this problem, we:
\begin{itemize}
\item Define the task of identifying persons killed by police, which is an instance
of cross-document entity-event extraction (\S \ref{s:task}).

\item Present a new dataset of web news articles collected throughout 2016 that describe possible fatal encounters with police officers (\S \ref{ss:newsdocs}).


\item Introduce, for the database update setting, a distant supervision model (\S\ref{s:model}) that incorporates feature-based logistic regression and convolutional neural network classifiers under a latent disjunction model.


\item  Demonstrate the approach's potential usefulness for practitioners: 
it outperforms two off-the-shelf event extractors (\S \ref{ss:offshelf})
and finds 39 persons not included in the Guardian's ``The Counted'' database of police fatalities as of January 1, 2017 (\S \ref{s:results}).  This constitutes
a promising first step, though performance needs to be improved for real-world usage.

\end{itemize}

\section{Related Work}

This task combines elements of information extraction,
including:
 \emph{event extraction} (a.k.a.\ \emph{semantic parsing}),
 identifying descriptions of events and their arguments from text,
 and cross-document \emph{relation extraction}, predicting semantic relations over entities.
 A fatality event indicates the killing of a particular person; 
we wish to specifically identify the names of fatality victims mentioned in text.
Thus our task could be viewed as unary relation extraction: for a given person mentioned in a corpus,
were they killed by a police officer?

Prior work in NLP has produced a number of event extraction systems,
trained on text data hand-labeled with a pre-specified ontology,
including ones that identify instances of killings 
\citep{Li2014,Das2014}.
Unfortunately, they
perform poorly on our task (\S\ref{ss:offshelf}),
so we develop a new method.

Since we do not have access to text specifically annotated for police killing events,
we instead turn to \emph{distant supervision}---inducing labels 
by aligning relation-entity entries  from a gold standard database
to their mentions in a corpus
\citep{Craven1999Bio,Mintz2009Distant,Bunescu2007Distant,Riedel2010Distant}.
Similar to this work,
\citet{Reschke2014} apply distant supervision to multi-slot,
template-based event extraction for airplane crashes; we focus on a simpler unary extraction setting with joint learning of a probabilistic model.
Other related work in the cross-document setting has examined joint inference
for relations, entities, and events \citep{Yao2010,Lee2012,Yang2015EventsTACL}.

Finally, other natural language processing efforts
have sought to extract social behavioral event databases 
from news, such as instances of
protests \citep{Hanna2017Protests},
gun violence \citep{Pavlick2016Guns},
and
international relations 
\citep{Schrodt1994,Schrodt2012Review,Boschee2013,OConnor2013IR,Gerrish2013Thesis}.
They can also be viewed as event database population tasks, with differing levels of semantic specificity
in the definition of ``event.''

\begin{table}[t]
\centering
{\footnotesize
\begin{tabular}{p{2.8cm}  p{1.5cm} p{1.5cm} }
\toprule
  \textbf{Knowledge base} & \textbf{Historical} & \textbf{Test }\\
\midrule
  FE incident dates & 
  Jan~2000 -- Aug~2016 
  & 
  Sep~2016 -- Dec~2016 
  \\
FE gold entities ($\mathcal{G}$) & 17,219 &  452 \\
\toprule
\textbf{News dataset} & \textbf{Train} & \textbf{Test }\\
\midrule
doc. dates  &
Jan~2016 -- Aug~2016
&  
 Sep~2016 -- Dec~2016 
 \\
total docs. ($\mathcal{D}$)& 866,199 &  347,160 \\
total \ ments.\  ($\mathcal{M}$)& 132,833 & 68,925\\
pos.\ ments. ($\mathcal{M^+}$) & 11,274 & 6,132 \\
total entities \ ($\mathcal{E}$)& 49,203 & 24,550 \\
pos.\ entities ($\mathcal{E^+}$) & 916 & 258 \\
\toprule
\end{tabular}
 }
\caption{ Data statistics for Fatal Encounters (FE) and scraped news documents.
$\mathcal{M}$ and $\mathcal{E}$ result from NER processing, 
while $\mathcal{E}^+$ results from matching textual named entities
against the gold-standard database $(\mathcal{G})$.
\label{t:data}}
\end{table}


\section{Task and Data}
\label{s:data}

\subsection{Cross-document entity-event extraction for police fatalties}
\label{s:task}
From a corpus of documents $\mathcal{D}$,
the task is to extract a list of candidate person names, $\mathcal{E}$, and for each  $e \in \mathcal{E}$ find
\begin{equation}
 P(y_e=1 \mid x_{\mathcal{M}(e)}). 
 \end{equation}
Here $y \in \{0, 1\}$ is the entity-level label where $y_e = 1$ means a person (entity) $e$ was killed by police; $x_{\mathcal{M}(e)}$ are the  sentences containing mentions $\mathcal{M}(e)$ of that person. 
A mention $i \in \mathcal{M}(e)$ is a token span in the corpus.
Most entities have multiple mentions;
a single sentence can contain multiple mentions of different entities.

\subsection{News documents}
\label{ss:newsdocs}
We download a collection of web news articles
by continually querying Google News\footnote{\url{https://news.google.com/}} 
throughout 2016
with lists of police keywords (i.e police, officer, cop etc.) 
and fatality-related keywords (i.e. kill, shot, murder etc.).
The keyword lists
were
constructed semi-automatically from cosine similarity lookups
from the \emph{word2vec}
pretrained word embeddings\footnote{\url{https://code.google.com/archive/p/word2vec/}}
in order to select a high-recall, broad set of keywords.
The search is restricted to what Google News defines as
a ``regional edition'' of ``United States (English)''
which seems to roughly restrict to U.S.\ news
though we 
anecdotally observed instances of news about events in the U.K. and other countries.
We apply a pipeline of text extraction, cleaning, and sentence de-duplication described in the appendix.

\subsection{Entity and mention extraction}
\label{ss:mentions}
We process all documents with
the open source \emph{spaCy} NLP package\footnote{Version 0.101.0, \url{https://spacy.io/}} 
to segment sentences, and extract entity mentions.
Mentions are token spans 
that (1)  were identified as ``persons'' by spaCy's
named entity recognizer,
and (2) have a (firstname, lastname) pair as analyzed by the
HAPNIS rule-based name parser,\footnote{\url{http://www.umiacs.umd.edu/~hal/HAPNIS/}}
which extracts, for example,
(\emph{John}, \emph{Doe}) from the string
\emph{Mr.\ John A.\ Doe Jr.}.\footnote{For both training and testing, we use a 
name matching assumption
that a (firstname, lastname) match indicates coreference between mentions,
and between a mention and a fatality database entity.
This limitation does affect a small number of instances---the test set database contains 
the unique names of 453 persons but only 451 unique (firstname, lastname) tuples---but relaxing it raises complex issues for future work, such as
how to evaluate whether a system correctly predicted two different fatality victims with the same name.}

To prepare sentence text for modeling, our preprocessor
collapses
the candidate mention span to a special TARGET symbol.
To prevent overfitting, other person names are mapped to a different PERSON symbol;
e.g.~``TARGET was killed in an encounter with police officer PERSON.''

There were initially 18,966,757 and 6,061,717 extracted mentions for the train and test periods respectively.
To improve precision and computational efficiency, we
filtered to sentences that contained at least one police keyword and one fatality keyword.
This filter reduced positive entity recall a moderate amount (from 0.68 to 0.57),
but removed 99\% of the mentions, resulting in the $|\mathcal{M}|$ counts in Table \ref{t:data}.\footnote{In preliminary experiments, training and testing an n-gram classifier ($\S\ref{sss:lr}$)
on the full mention dataset without keyword filtering resulted in a worse AUPRC than after the filter.}

Other preprocessing steps included heuristics for extraction and name cleanups and are detailed in the appendix.

\section{Models}
\label{s:model}

\begin{table}[t]
\footnotesize
\centering
\begin{tabular}{p{2.6cm} p{1cm} p{1.9cm} p{1cm}}
\toprule
& $x$ & $z$ & $y$
 \\
\midrule
``Hard" training & observed & fixed (distantly labeled) & observed \\
``Soft" (EM) training & observed & latent & observed \\
Testing & observed & latent & latent \\
\toprule 
\end{tabular} 
\caption{Training and testing settings for mention sentences $x$, mention labels $z$, and entity labels $y$. 
\label{t:modelvar}
}
\end{table}

Our goal is to classify entities as to whether they have been killed by police (\S\ref{ss:class}). 
Since we do not have gold-standard labels to train our model,
we turn to \emph{distant supervision} \citep{Craven1999Bio,Mintz2009Distant},
which
heuristically aligns facts in a knowledge base to text in a corpus
to impute positive mention-level labels for supervised learning.
Previous work typically examines distant supervision in the context of binary relation extraction
\citep{Bunescu2007Distant,Riedel2010Distant,Hoffmann2011Distant},
but we are concerned with the unary predicate ``person was killed by police.''
As our gold standard knowledge base ($\mathcal{G}$), we use Fatal Encounters' (FE) publicly available dataset:
around 18,000 entries of victim's name, age, gender and race as well as location, cause and date of death.
(We use a version of the FE database downloaded Feb.~27, 2017.)
We compare two different distant supervision training paradigms (Table~\ref{t:modelvar}):
``hard'' label training (\S\ref{ss:hard}) 
and ``soft" EM-based training (\S\ref{ss:soft}).
This section also details mention-level models (\S\ref{sss:lr},\S\ref{sss:cnn}) and evaluation (\S\ref{ss:eval}).

\subsection{Approach: Latent disjunction model} 
\label{ss:class}
Our discriminative model is built on mention-level probabilistic classifiers. 
Recall a single entity will have one or more mentions 
(i.e.\ the same name occurs in multiple sentences in our corpus).
For a given mention $i$ in sentence $x_i$, 
our model predicts whether the person is described as having been killed by police, $z_i = 1$,
with a binary logistic model,
\begin{equation}\label{eq:z_given_x}
\hspace{-0.1in}P(z_i=1 \mid x_i) = \sigma(\beta\transpose f_{\gamma}(x_i)).
\end{equation}
We experiment with both logistic regression (\S\ref{sss:lr})
and convolutional neural networks (\S\ref{sss:cnn})
for this component, which use logistic regression weights $\beta$ and feature extractor parameters $\gamma$.
Then we must somehow aggregate mention-level decisions to determine entity labels $y_e$.\footnote{An alternative approach
is to aggregate features across mentions into an entity-level feature vector
\citep{Mintz2009Distant,Riedel2010Distant}; 
but here we opt to directly model 
at the mention level, which can use contextual information.}
If a human reader were to observe at least one sentence that states a person was killed by police, they would infer that person was killed by police.  Therefore we aggregate an entity's mention-level labels
with a 
deterministic disjunction:
\begin{equation}
 P(y_e =1\mid z_{\mathcal{M}(e)} ) = 1\left\{ \vee_{i \in \mathcal{M}(e)} \ z_i \right\}.
 \label{e:yedisj}
\end{equation}
At test time, $z_i$ is latent.  Therefore the correct inference for an entity is to marginalize out the model's uncertainty over $z_i$:
 \begin{align}
&P(y_e = 1 | x_{\mathcal{M}(e)} ) 
=  1 - P(y_e = 0 | x_{\mathcal{M}(e)} )
 \\
&\hspace{0.3in}=
 1 -  P(z_{\mathcal{M}(e)}=\vec{0} \mid x_{\mathcal{M}(e)})
 \\
&\hspace{0.3in} =
 1- \prod_{i \in \mathcal{M}(e)} (1-P(z_i=1\mid x_i)) \label{e:noisyor}.
 \end{align}
Eq.~\ref{e:noisyor} is the \emph{noisyor}
formula \citep{Pearl1988,Craven1999Bio}. 
Procedurally, it counts strong probabilistic predictions as evidence, 
but can
also incorporate a large number of weaker signals as positive evidence as well.\footnote{In early experiments,
we experimented with other, more ad-hoc aggregation rules with a ``hard''-trained model.
The maximum and arithmetic mean functions performed worse than \emph{noisyor},
giving credence to the disjunction model.
The sum rule ($\sum_i P(z_i=1 \mid x_i)$)
had similar ranking performance as \emph{noisyor}---perhaps because it too can use weak signals, unlike mean or max---though it does not yield proper probabilities between 0 and 1.}


In order to train these classifiers, 
we need mention-level labels ($z_i$) which we impute 
via two different distant supervision labeling methods: ``hard'' and ``soft."

\subsection{``Hard" distant label training}
\label{ss:hard}
In ``hard'' distant labeling, labels for mentions in the training data are heuristically imputed and directly used for training.
We use two labeling rules. First, \textbf{name-only:}
 \begin{equation}
\begin{split}
z_i = 1 \text{\ if\ } \exists e \in \mathcal{G}^{(train)}: 
 \text{name}(i) = \text{name}(e).
 \end{split}
 \end{equation}
 This is the direct unary predicate analogue of 
\citet{Mintz2009Distant}'s \emph{distant supervision assumption},
which assumes every mention of a gold-positive entity exhibits a
description of a police killing.

This assumption is not correct. We manually analyze a sample of positive mentions and find
36 out of 100 name-only sentences did not express a police fatality event---for example, sentences contain commentary, or describe killings not by police.
This is similar to the precision for distant supervision of binary relations found by \citet{Riedel2010Distant},
who reported
10--38\% of sentences did not express the relation in question.

Our higher precision rule, \textbf{name-and-location}, leverages the fact that the location of the fatality is also
 in the Fatal Encounters database and requires both to be present:
\begin{equation}
\begin{split}
z_i = 1 \text{\ if\ } \exists e \in \mathcal{G}^{(train)}: \hspace{1in}\\
\text{name}(i)=\text{name}(e)
 \text{ and } \text{location}(e) \in x_i.
 \end{split}
\end{equation}
We use this rule for training since precision is slightly better, although there is still a considerable level of noise.
 
 \subsection{``Soft" (EM) joint training}
 \label{ss:soft}
 
 At training time, the \emph{distant supervision assumption} used in ``hard'' label training is flawed:
many positively-labeled mentions are in sentences that do not assert the person was killed by a police officer.
Alternatively, at training time we can treat $z_i$ as a latent variable and assume, as our model states,
that \emph{at least one} of the mentions asserts the fatality event,
but leave uncertainty over which mention (or multiple mentions)
conveys this information.
This corresponds to multiple instance learning (MIL; \citet{Dietterich1997MIL})
which has been applied to distantly supervised relation extraction
by enforcing the  \emph{at least one} constraint at training time
\citep{Bunescu2007Distant,Riedel2010Distant,Hoffmann2011Distant,Surdeanu2012Distant,Ritter2013Missing}.
Our approach differs by using exact
marginal posterior inference for the E-step.

\label{ss:em_latent}
With $z_i$ as latent, the model can be trained with the EM algorithm
\citep{Dempster1977EM}.
We initialize the model by training
on the ``hard" distant labels (\S\ref{ss:hard}),
and then learn improved parameters by alternating E- and M-steps.

The \textbf{E-step} requires calculating the marginal posterior probability for each $z_i$,
\begin{equation}
q(z_i) := P(z_i \mid x_{\mathcal{M}(e_i)}, y_{e_i}). \label{eq:qzi}
\end{equation}
\noindent
This corresponds to calculating the posterior probability of a disjunct, given knowledge of the output of the disjunction, and prior probabilities of all disjuncts (given by the mention-level classifier).

Since $P(z \mid x,y) = P(z,y \mid x)/P(y\mid x)$,
\begin{align} \label{e:disjpost}
q(z_i=1) &=
 \frac{P(z_i =1, y_{e_i} = 1 | x_{\mathcal{M}(e_i)})}{P(y_{e_i} =1 | x_{\mathcal{M}(e_i)})}.
\end{align}
The numerator simplifies to the mention prediction $P(z_i =1 \mid x_i)$
and the denominator is the entity-level \emph{noisyor} probability (Eq.~\ref{e:noisyor}).
This has the effect of taking the classifier's predicted probability and increasing it slightly
(since Eq.~\ref{e:disjpost}'s denominator is no greater than 1);
thus the disjunction constraint implies a soft positive labeling.
In the case of a negative entity with $y_e=0$,
the disjunction constraint implies all $z_{\mathcal{M}(e)}$ stay clamped to 0
as in the ``hard" label training method.

The $q(z_i)$ posterior weights are then used for the
\textbf{M-step}'s expected log-likelihood objective:
\begin{equation}
\max_\theta \sum_i \sum_{z \in \{0,1\}} q(z_i=z) \log P_\theta(z_i=z \mid x_i).\label{eq:mstep}
\end{equation}
This objective (plus regularization) is maximized with gradient ascent as before.

\begin{figure}[t]\hspace{0.08in}
\includegraphics[width=8cm]{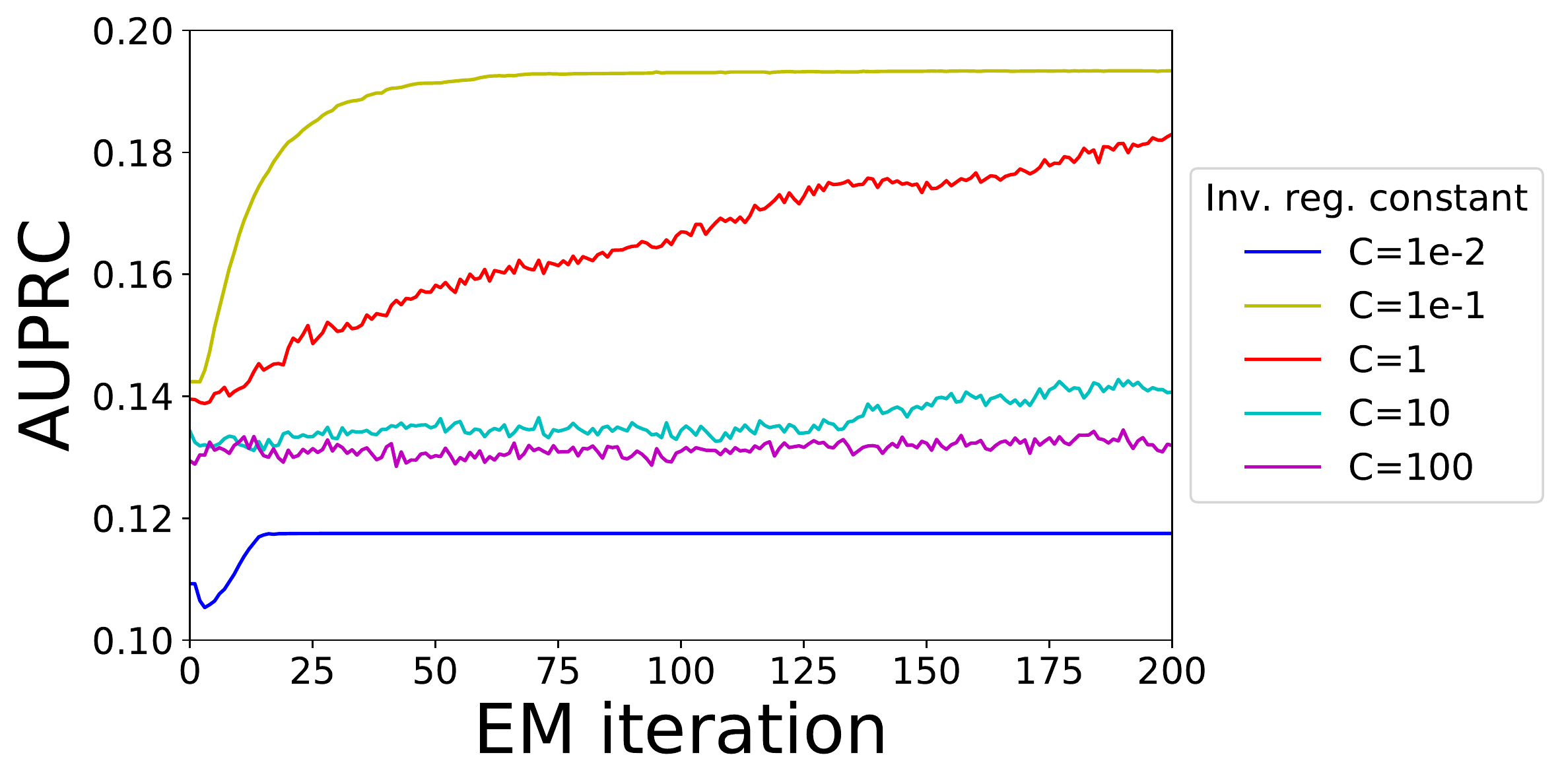}
\caption{For soft-LR (EM), area under precision recall curve (AUPRC) results 
on the test set
during training,
for different inverse regularization values
($C$, the parameters' prior variance).
\label{f:tunec}}
\end{figure}

This approach can be applied to any mention-level probabilistic model; we explore
two in the next sections.

 \subsection{Feature-based logistic regression}
\label{sss:lr}
\begin{table}[t]
{\footnotesize
\centering
\begin{tabular}{p{0.2cm} p{6.5cm}}
\toprule
& Features \\
\midrule
$D1$ & length 3 dependency paths that include TARGET: word, POS, dep.\ label
\\
$D2$ & length 3 dependency paths that include TARGET:  word and dep.\ label
\\
$D3$ & length 3 dependency paths that include TARGET:  word and POS
\\
$D4$ & all length 2  dependency paths with word, POS, dep. labels
\\
\midrule
$N1$ & n-grams length 1, 2, 3
\\
$N2$ & n-grams  length 1, 2, 3 plus POS tags
\\
$N3$ & n-grams length 1, 2, 3 plus directionality and position from TARGET
\\
$N4$ & concatenated POS tags of 5-word window centered on TARGET
\\
$N5$ & word and POS tags for 5-word window centered on TARGET
\\
\toprule 
\end{tabular} 
}
\caption{Feature templates for logistic regression grouped into syntactic dependencies $(D)$ and N-gram 
$(N)$ features. 
\label{t:feats}
}
\end{table}

We construct hand-crafted features for regularized logistic regression (LR)
(Table \ref{t:feats}), designed to be broadly similar to the n-gram and syntactic 
dependency features used in previous work on feature-based
semantic parsing (e.g.\ \citet{Das2014,Thomson2014SemEval}).
We use randomized feature hashing \citep{Weinberger2009Hashing}
to efficiently represent features in 450,000 dimensions,
which achieved similar performance as an explicit feature representation.
The logistic regression weights ($\beta$ in Eq.\ \ref{eq:z_given_x})
are learned with \emph{scikit-learn} \citep{scikit-learn}.\footnote{With \emph{FeatureHasher}, L2 regularization, `lbfgs' solver,
and inverse strength $C=0.1$, tuned on a development dataset in ``hard'' training; 
for EM training the same regularization strength performs best.}
For EM (soft-LR) training, 
the test set's area under the precision recall curve converges after 96 iterations (Fig.\ \ref{f:tunec}).

\subsection{Convolutional neural network}
\label{sss:cnn}
We also train a convolutional neural network (CNN) classifier,
which uses word embeddings and their nonlinear compositions
to potentially generalize better than sparse lexical and n-gram features.
CNNs have been shown useful for
sentence-level classification tasks
\cite{Kim2014ConvolutionalNN,zhang2015sensitivity}, relation classification \citep{Zeng2014} and, 
similar to this setting,
event detection \citep{Nguyen2015}.
We use \citet{Kim2014ConvolutionalNN}'s open-source CNN implementation,\footnote{\url{https://github.com/yoonkim/CNN_sentence}}
where a logistic function makes the final mention prediction
based on max-pooled values
from convolutional layers of three different filter sizes, whose parameters are learned ($\gamma$ in Eq.~\ref{eq:z_given_x}).
We use pretrained word embeddings for initialization,\footnote{From the same \emph{word2vec} embeddings 
used in \S\ref{s:data}.} and update them during training.
We also add two special vectors for the TARGET and PERSON symbols, initialized randomly.\footnote{Training proceeds with ADADELTA \citep{Zeiler2012ADADELTAAA}. We tested several different settings of dropout and L2 regularization hyperparameters on a development set, but found mixed results, so used their default values.} 

For training, we perform stochastic gradient descent for the negative expected log-likelihood
(Eq.\ \ref{eq:mstep}) by sampling with replacement fifty mention-label pairs for each minibatch,
choosing each $(i,k) \in \mathcal{M} \times \{0,1\}$
with probability proportional to $q(z_i=k)$.
This strategy attains the same expected gradient as the overall objective.
We use ``epoch'' to refer to training on 265,700 examples (approx.\ twice the number of mentions).
Unlike EM for logistic regression, we do not run gradient descent to convergence,
instead applying an E-step every two epochs to update $q$;
this approach is related to incremental and online variants of EM
\citep{Neal1998EM,Liang2009EM},
and is justified since both SGD and E-steps improve the evidence lower bound
(ELBO).  
It is also similar to \newcite{Salakhutdinov2003ECG}'s expectation gradient method;
their analysis implies the gradient calculated immediately after an E-step
is in fact the gradient for the
marginal log-likelihood.
We are not aware of recent work that uses EM to train latent-variable
neural network models, though this combination has been explored
(e.g.\ \citet{Jordan1994HME})

\subsection{Evaluation}
\label{ss:eval}

\begin{figure}[t]
\includegraphics[width=8cm]{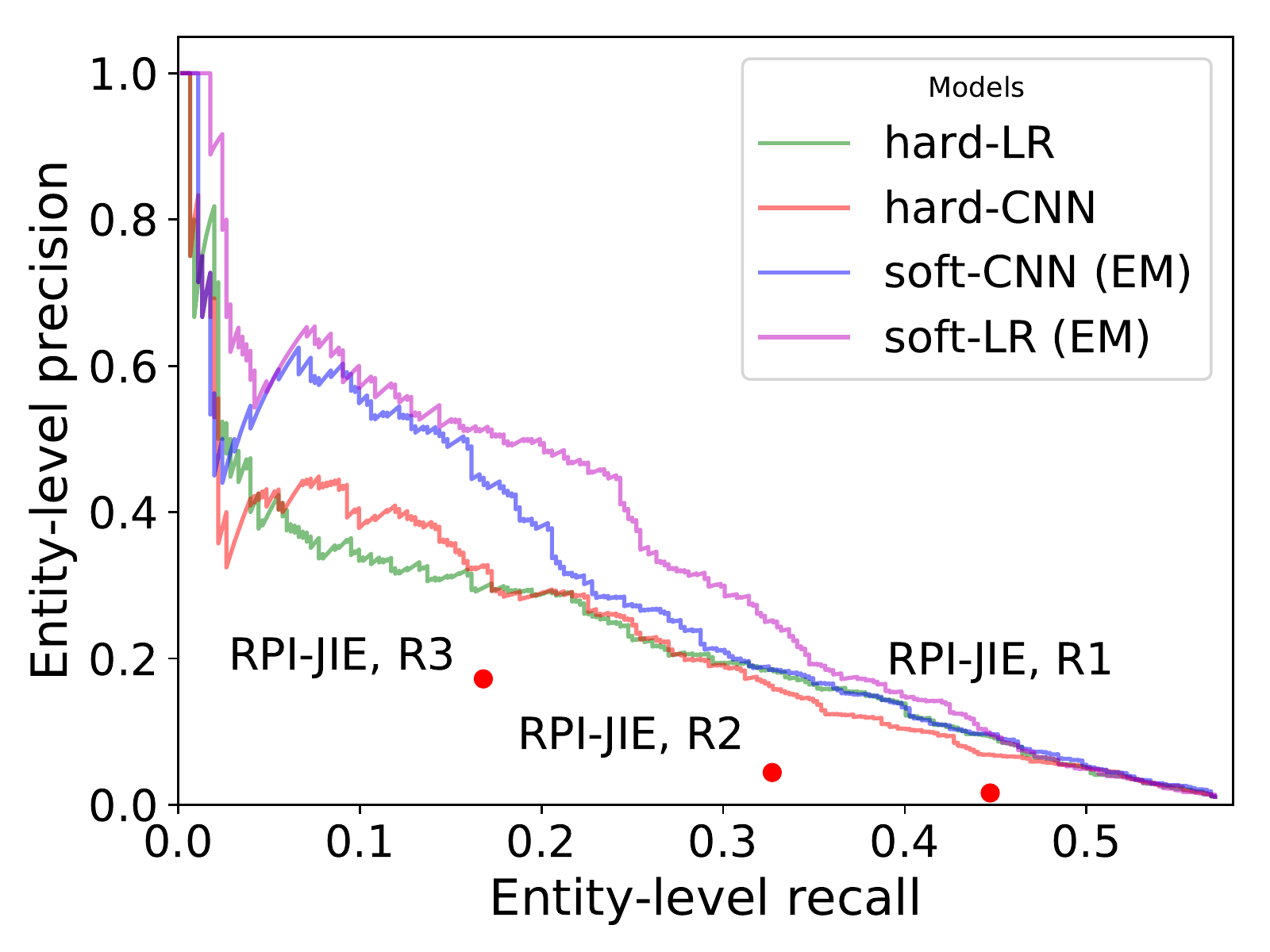}
\caption{Precision-recall curves for the given models. 
\label{f:prrec}}
\end{figure}

On documents from the test period (Sept--Dec\ 2016), our models
predict entity-level labels $P(y_e = 1 \mid x_{\mathcal{M}(e)} )$ (Eq.~\ref{e:noisyor}),
and we wish to evaluate whether retrieved entities are listed in
Fatal Encounters as being killed during Sept--Dec\ 2016.
We rank entities by predicted probabilities to construct a precision-recall curve
(Fig. \ref{f:prrec}, Table \ref{t:results}).
Area under the precision-recall curve (AUPRC) is calculated with a trapezoidal rule;
F1 scores are shown for convenient comparison to non-ranking approaches (\S\ref{ss:offshelf}).


\begin{figure}[t]
\includegraphics[width=8cm]{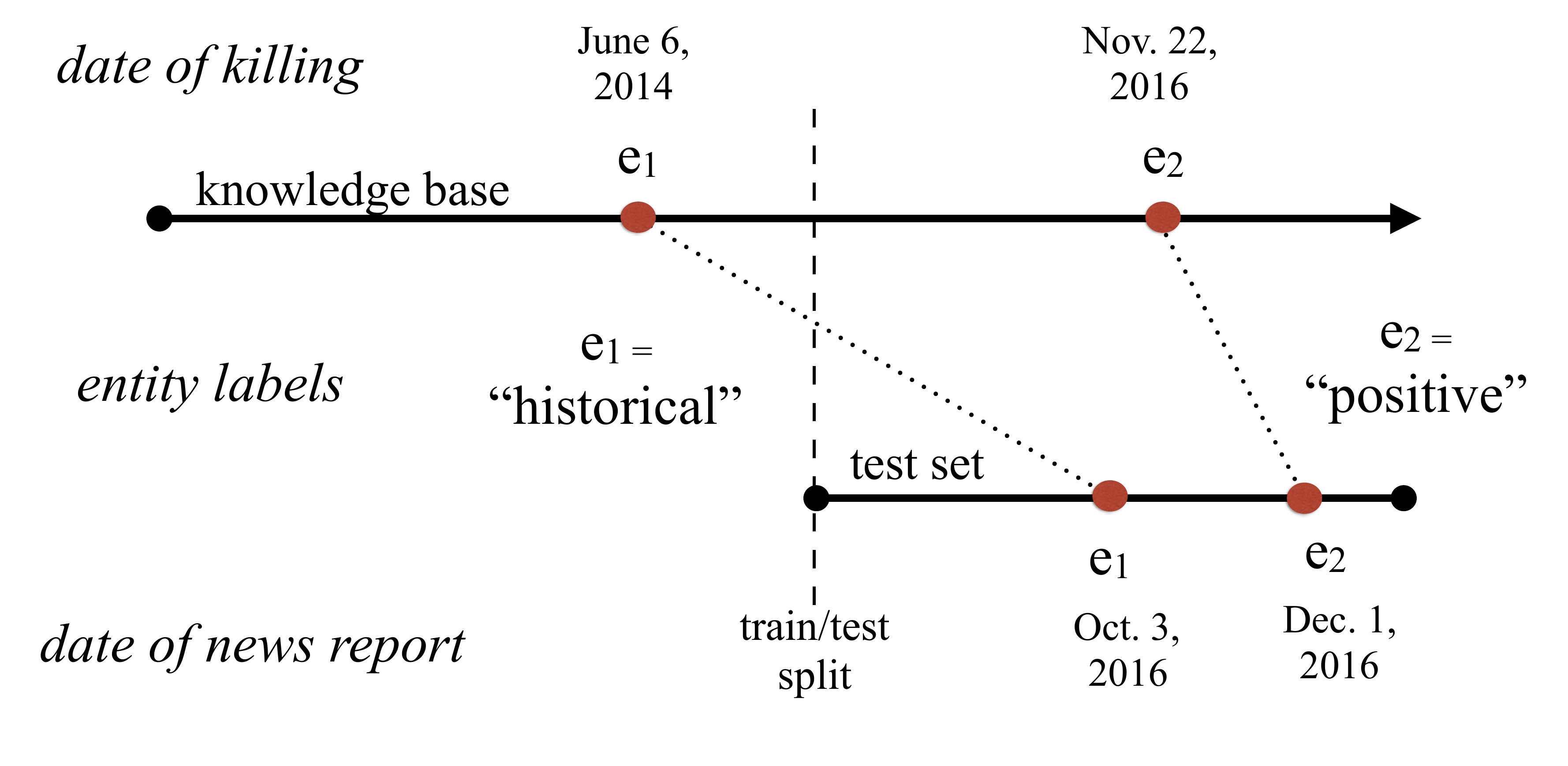}
\caption{At test time, there are matches between the knowledge base and the news reports both for persons
killed during the test period (``positive") and persons killed before it (``historical'').  Historical cases
are excluded from evaluation.
\label{f:hist}}
\end{figure}

\begin{figure}[t]
\centering
\includegraphics[width=6cm]{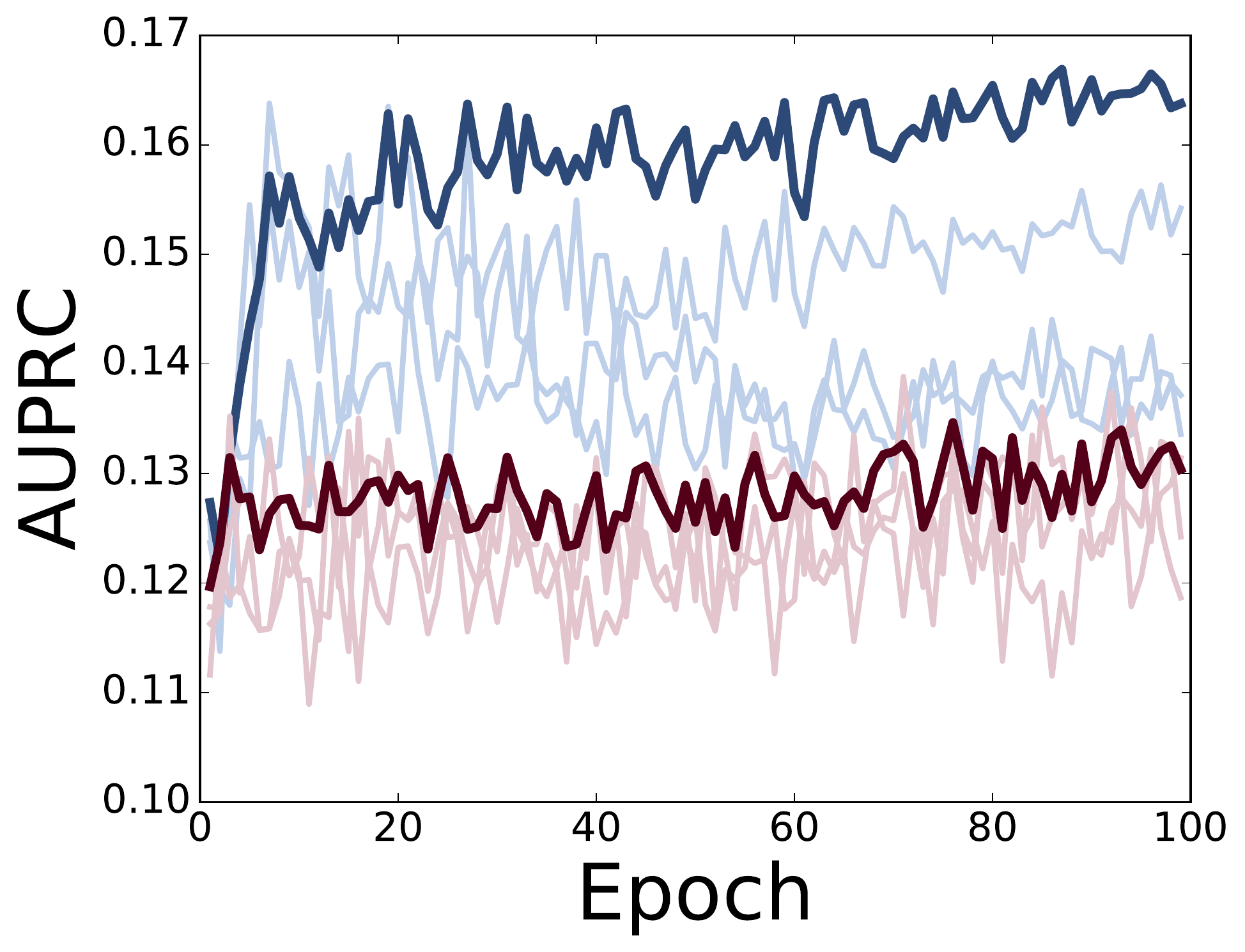}
\caption{
Test set AUPRC for three runs of 
soft-CNN (EM) (\textbf{\textcolor[RGB]{45,74,117}{blue}}, higher in graph),
and
hard-CNN (\textbf{\textcolor[RGB]{83,2,25}{red}}, lower in graph).
Darker lines show performance of averaged predictions. 
\label{f:cnn}}
\end{figure}

\begin{table}[t]
\centering
\begin{tabular}{  l  r  r }
\toprule
Model & AUPRC & F1 \\
\midrule
hard-LR, dep.\ feats. & 0.117 & 0.229 \\
hard-LR, n-gram\ feats. & 0.134 & 0.257 \\
hard-LR, all feats. & 0.142 & 0.266 \\
hard-CNN & 0.130 & 0.252 \\
\midrule
soft-CNN (EM) & 0.164 & 0.267 \\
\textbf{soft-LR (EM)} &  \textbf{0.193} & \textbf{0.316} \\
\toprule
Data upper bound (\S\ref{ss:eval}) & 0.57 & 0.73
\\
\toprule
\end{tabular}
\caption{Area under precision-recall curve (AUPRC) and F1 (its maximum value from the PR curve)
for entity prediction on the test set. 
\label{t:results}}
\end{table}

\textbf{Excluding historical fatalities:}
Our model gives strong positive predictions for many people who were killed by police before the
test period (i.e.\ before Sept 2016), when news articles contain discussion of historical
police killings. We exclude these entities from evaluation,
since we want to simulate an update to a fatality database (Fig \ref{f:hist}).
Our test dataset contains 1,148 such historical entities.

\textbf{Data upper bound:} 
Of the 452 gold entities in the FE database at test time, our news corpus only contained 258 (Table \ref{t:data}), hence the data upper bound of 0.57 recall, which also gives an upper bound of 0.57 on AUPRC. 
This is mostly a limitation of our news corpus; though we collect hundreds of thousands
of news articles, it turns out
Google News only accesses a subset of relevant web news, as opposed
to more comprehensive data sources
manually reviewed by Fatal Encounters' human experts.
We still believe our dataset is large enough
to be realistic for developing better methods, and expect the same approaches could be applied to a more comprehensive news corpus.

\section{Off-the-shelf event extraction baselines}
\label{ss:offshelf}

\begin{table}[t]
\footnotesize
\centering
\begin{tabular}{l l l l l }
\toprule
 & Rule & Prec. & Recall & F1
 \\
\midrule
SEMAFOR & R1 & 0.011 &0.436 & 0.022
\\
& R2 & 0.031 &  0.162 & 0.051
\\ 
& R3 & 0.098 & 0.009 & 0.016
\\
\midrule 
RPI-JIE & R1 & 0.016 & 0.447 & 0.030
\\
& R2 & 0.044 & 0.327 & 0.078
\\ 
& R3 & 0.172 &  0.168 & \textbf{0.170}
\\
\toprule
Data upper bound (\S\ref{ss:eval}) & & 1.0 & 0.57 & 0.73
\\
\toprule 
\end{tabular} 
\caption{Precision, recall, and F1 scores for test data using event extractors SEMAFOR and RPI-JIE and rules R1-R3 described below. 
\label{t:extract}
}
\end{table}
From a practitioner's perspective, a natural first approach to this task would be to run the corpus of police fatality documents through pre-trained, ``off-the-shelf" event extractor systems that could identify killing events.  
In modern NLP research,
a major paradigm for event extraction is to formulate a hand-crafted ontology of event classes,
annotate a small corpus, and craft supervised learning systems to predict event parses of documents.

We evaluate two freely available, off-the-shelf event extractors that were developed
under this paradigm:
SEMAFOR \citep{Das2014},
and the RPI Joint Information Extraction System (RPI-JIE)  \citep{Li2014},
which output semantic structures following
the FrameNet \citep{Fillmore2003}
and ACE \citep{doddington2004automatic}
 event ontologies, respectively.\footnote{Many other annotated datasets encode similar event structures in text,
but with lighter ontologies where event classes directly correspond with lexical items---including
PropBank, Prague Treebank, DELPHI-IN MRS,
and Abstract Meaning Representation
\citep{Kingsbury2002Propbank,Hajic2012Prague,Oepen2014Semeval,Banarescu2013AMR}.
We assume such systems are too narrow for our purposes,
since we need an extraction system to handle different trigger constructions
 like ``killed'' versus ``shot dead.''}
\citet{Pavlick2016Guns} use RPI-JIE to identify instances of gun violence.

For each mention
$ i \in \mathcal{M}$ we use SEMAFOR and RPI-JIE to extract event tuples of the form  $t_i = \text{(event type, agent, patient)}$ from the sentence $x_i$.
We want the system to detect (1) killing events, where (2) the killed person is the target mention $i$,
and (3) the person who killed them is a police officer.  
We implement a small progression of these 
neo-Davidsonian \citep{Parsons1990Events}
conjuncts with rules to classify $z_i=1$ if:\footnote{For SEMAFOR, we use the FrameNet `Killing' frame with frame elements
`Victim' and `Killer'. For RPI-JIE, we use the ACE `life/die' event type/subtype
with roles `victim' and `agent'. 
SEMAFOR defines a token span for every argument; RPI-JIE/ACE defines two spans, both a head word and entity extent; we use the entity extent.  
SEMAFOR only predicts spans as event arguments, while
RPI-JIE also predicts entities as event arguments, where each entity has a within-text coreference chain over one or more mentions;
since we only use single sentences, these chains tend to be small,
though they do sometimes resolve pronouns. 
For determining R2 and R3, we allow a match on any of an entity's extents from any of its mentions.}

\begin{itemizesquish}
\item
\textbf{(R1)} the event type is `kill.'

\item
\textbf{(R2)} R1 holds and the patient token span contains $e_i$.

\item
\textbf{(R3)} R2 holds and the agent token span contains a police keyword.
\end{itemizesquish}

\noindent
As in \S\ref{ss:class} (Eq. \ref{e:yedisj}), we aggregate mention-level $z_i$ predictions to obtain entity-level predictions
with a deterministic OR of $z_{\mathcal{M}(e)}$.

\begin{table*}[t]
\centering
\small
 \begin{tabular}{p{2.5cm} p{1.0cm}  p{12.0cm} }
 \toprule
  entity ($e$) & ment.($i$) prob. & ment. text ($x_i$) \\
\midrule
\multirow{2}{*}
  {\pbox{2.0cm}{\textbf{Keith Scott} \\ (true pos) 
  }}
  & 0.98 
  & Charlotte protests Charlotte's Mayor Jennifer Roberts speaks to reporters the morning after protests against the police shooting of \textbf{Keith Scott}, in Charlotte, North Carolina .  \\
\midrule
  \multirow{2}{*}
  {\pbox{2.0cm}{\textbf{Terence Crutcher} \\ (true pos) 
  }}
  & 0.96 
  & Tulsa Police Department released video footage Monday, Sept. 19, 2016, showing white Tulsa police officer Betty Shelby fatally shooting \textbf{Terence Crutcher,} 40, a black man police later determined was unarmed.
  \\
  \midrule
   \multirow{2}{*}
  {\pbox{2.3cm}{\textbf{Mark Duggan} \\ (false pos)
  }}
  & 0.97
  & The fatal shooting of \textbf{Mark Duggan} by police led to some of the worst riots in England's recent history.\\
\midrule
  \multirow{2}{*}
  {\pbox{2.3cm}{\textbf{Logan Clarke} \\ (false pos)
  }}
  & 0.92 
  & \textbf{Logan Clarke} was shot by a campus police officer after waving kitchen knives at fellow students outside the cafeteria at Hug High School in Reno, Nevada, on December 7. \\
\toprule 
  \end{tabular}
  \caption{Example of highly ranked entities, with selected mention predictions and text. 
  \label{t:printout}}
\end{table*}

RPI-JIE under the full R3 system performs best,
though all results are relatively poor (Table~\ref{t:extract}).
Part of this is due to inherent difficulty of the task, though
our task-specific model still outperforms (Table \ref{t:results}).
We suspect a major issue is that these systems heavily rely on their annotated training sets
and may have significant performance loss on new domains, or messy text extracted from web news,
suggesting domain transfer for future work.

\section{Results and discussion}
\label{s:results}

\textbf{Significance testing:}
We would like to test robustness of performance results to the finite datasets
with bootstrap testing \citep{BergKirkpatrick2012Significance},
which can accomodate performence metrics like
AUPRC.  It is not clear what the appropriate unit of resampling should be---for
example, parsing and machine translation research in NLP often 
resamples sentences, which is inappropriate for our setting.
We elect to resample documents in the test set,
simulating variability in the generation and retrieval
of news articles.  Standard errors for one model's AUPRC and F1
are in the range 0.004--0.008 and 0.008--0.010 respectively;
we also note pairwise significance test results.
See appendix for details.

\textbf{Overall performance:} 
Our results indicate our model is better than existing computational methods
methods to extract names of people killed by police,
by comparing to F1 scores of off-the-shelf extractors
(Table \ref{t:results} vs.\ Table \ref{t:extract}; differences are statistically significant).

We also compare entities extracted from our test dataset to the Guardian's ``The Counted" database
of U.S.\ police killings during the span of the test period (Sept.--Dec., 2016),\footnote{\url{https://www.theguardian.com/us-news/series/counted-us-police-killings}, downloaded Jan.~1, 2017.}
and found 39 persons they did not include in the database,
but who were in fact killed by police.  This implies our approach could
augment journalistic collection efforts.
Additionally, our model could help practitioners by presenting them with sentence-level
information in the form of Table \ref{t:printout};
we hope this could decrease the amount of time and emotional toll
required to maintain real-time updates of police fatality databases.

\textbf{CNN:} 
Model predictions were relatively unstable during the training process.
Despite the fact that EM's evidence lower bound objective ($H(Q)+E_Q[\log P(Z,Y|X)]$)
converged fairly well on the training set, 
test set AUPRC substantially fluctuated
as much as 2\% between epochs,
 and also between three different random initializations for training
(Fig.~\ref{f:cnn}).
We conducted these multiple runs initially to check for variability, then used them to construct a basic ensemble:
we averaged the three models'
mention-level predictions
before applying \emph{noisyor} aggregation. 
This outperformed the individual models---especially for EM training---and showed less fluctuation in AUPRC, which made it easier to detect convergence.
Reported performance numbers in Table \ref{t:results} are with the average of all three runs from the final epoch of training.


\textbf{LR vs.\ CNN:}
After feature ablation we found that hard-CNN and hard-LR with n-gram features (N1-N5) had comparable AUPRC values (Table \ref{t:results}). 
But adding dependency features (D1-D4) caused the logistic regression models 
to outperform the
neural networks (albeit with bare significance: $p=0.046$).
We hypothesize these dependency features capture longer-distance semantic relationships between the entity, fatality trigger word, and police officer, which short n-grams cannot.  Moving to sequence or graph LSTMs may better capture such dependencies.

\textbf{Soft (EM) training:} Using the EM algorithm gives substantially
better performance: for the CNN, AUC improves from 0.130 to 0.164,
and for LR, from 0.142 to 0.193.  (Both improvements are statistically significant.)
Logistic regression with EM training is the most accurate model.
Examining the precision-recall curves (Fig.~\ref{f:prrec}),
many of the gains are in the higher confidence predictions (left side of figure).
In fact, the soft EM model makes fewer strongly positive predictions:
for example,
hard-LR predicts $y_e=1$ with more than 99\% confidence
for 170 out of 24,550 test set entities,
but soft-LR does so for only 24.
This makes sense given that the hard-LR model at training time
assumes that many more positive entity mentions are evidence of a killing
than they are in reality (\S\ref{ss:hard}).


\textbf{Manual analysis:} Manual analysis of false positives indicates misspellings or mismatches of names, police fatalities outside of the U.S., people who were shot by police but not killed, and names of police officers who were killed are common false positive errors (see detailed table in the appendix). 
This suggests many prediction errors are from ambiguous or challenging cases.\footnote{We attempted to correct non-U.S. false positive errors by using 
CLAVIN, an open-source country identifier, but this significantly hurt recall.}

\textbf{Future work:}
While we have made progress on this application, more work is necessary
for accuracy to be high enough to be useful for practitioners.  Our model allows for the use of
mention-level semantic parsing models; systems with explicit trigger/agent/patient representations,
more like traditional event extraction systems, may be useful, as would more sophisticated neural network models, or attention models as an alternative to disjunction aggregation \citep{Lin2016Distant}.

One goal is to use our model as part of a semi-automatic system, where people manually review a ranked list of entity suggestions. In this case, it is more important to focus on improving recall---specifically,
improving precision at high-recall points on the precision-recall curve.
Our best models, by contrast, tend to improve precision at lower-recall points on the curve.
Higher recall may be possible through cost-sensitive training
(e.g.\ \citet{Gimpel2010Softmax}) and using features from 
beyond single sentences within the document.

Furthermore, our dataset could be used to contribute to communication studies,
by exploring research questions about
the dynamics of media attention (for example, the effect of race and geography on coverage of police killings),
and discussions of historical killings in news---for example, many articles in 2016 discussed
Michael Brown's 2014 death in Ferguson, Missouri. 
Improving NLP analysis of historical events
would also be useful for 
the event extraction task itself,
by delineating between recent events that require a database update, versus historical events
that appear as ``noise'' from the perspective of the database update task.
Finally, it may also be possible to adapt our model to extract other types
of social behavior events.

\ifemnlpfinal
\section*{Acknowledgments}
This work was partially supported
by the Amazon Web Services (AWS) Cloud Credits for Research program.
Thanks to D.\ Brian Burghart for advice on police fatalities tracking,
and to David Belanger, Trapit Bansal, Patrick Verga, Rajarshi Das, and Taylor Berg-Kirkpatrick for feedback.
\fi

\bibliographystyle{emnlp_natbib}
\bibliography{brenocon,abe,katie}

\forarxiv{
\appendix
\section*{Appendix}
\section{Document retrieval from Google News}

Our news dataset is created using documents gathered via Google News.  Specifically, we issued search queries to Google News\footnote{\url{https://news.google.com/}} United States (English) regional edition
throughout 2016.
Our scraper issued queries with terms from two lists:
(1) a list of 22 words closely related to police officers
and (2) a list of 21 words closely related to killing.
These lists were semi-automatically constructed by looking up the nearest neighbors of ``police'' and ``kill'' (by cosine distance)
from Google's public release of \emph{word2vec} vectors pretrained
on a very large (proprietary) Google News corpus,\footnote{\url{https://code.google.com/archive/p/word2vec/}}
and then manually excluding a small number of misspelled words or redundant capitalizations (e.g.\ ``Police'' and ``police'').

Our list of {police words} includes: police, officer, officers,  cop, cops, detective, sheriff, policeman, policemen, constable, patrolman, sergeant, detectives, patrolmen, policewoman,  constables, trooper, troopers, sergeants, lieutenant, deputies, deputy. 

Our list of {kill words} includes: kill, kills, killing, killings, killed, shot, shots, shoot, shoots, shooting, murder, murders, murdered, beat, beats, beating, beaten, fatal, homicide, homicides.

We construct one word queries using single terms drawn from one of the two lists, as well as two-word queries which consist of one word drawn from each list (e.g. ``police shoot" or ``cops gunfire"),
yielding 505 different queries (22$\times$21 + 22 + 21), each of which was queried approximately once per hour throughout 2016.\footnote{We also collected data during part of 2015; the volume of search results varied over time due to changes internal to Google News.  After the first few weeks in 2016, the volume was fairly constant.}
This yielded a list of recent results matching the query;
the scraper downloaded documents whose URL it had not seen before,
eventually collecting 1,162,300 web pages (approx.~3000 per day).

\section{Document preprocessing}

\begin{table}[t]
\centering
\small
\begin{tabular}{l l l l}
\toprule
rank & name & positive & analysis 
\\
\midrule
1 &  \textbf{Keith Scott} & \textbf{true} &\\
2 & \textbf{Terence Crutcher} & \textbf{true} &\\
3 &  Alfred Olango & true &\\
4 &  Deborah Danner & true & \\
5  & Carnell Snell& true &\\
6 &  Kajuan Raye& true & \\
7 & Terrence Sterling & true &\\
8 & Francisco Serna & true &\\
9 &Sam DuBose& false  & name mismatch \\
10 & Michael Vance & true& \\
11 &Tyre King & true& \\
12  &Joshua Beal& true & \\
13  & Trayvon Martin&  false & killed, not by police \\
14  &\textbf{Mark Duggan}&  \textbf{false} & \textbf{non-US} \\
15  & Kirk Figueroa& true&\\
16  & Anis Amri&  false & non-US\\
17  &\textbf{Logan Clarke}&  \textbf{false} & \textbf{shot not killed}\\
18  & Craig McDougall&  false & non-US\\
19  & Frank Clark &true & \\
20 & Benjamin Marconi &  false & name of officer\\
\toprule
\end{tabular}
\caption{Top 20 entity predictions given by soft-LR (excluding historical entities) evaluated as ``true" or ``false" based on matching the gold knowledge base.
False positives were manually analyzed. 
See Table 7 in the main paper for more detailed information regarding bold-faced entities. 
\label{f:top20}
}
\end{table}

Once documents are downloaded from URLs collected via Google news queries, we apply text extraction with the Lynx browser\footnote{Version 2.8} to extract text from HTML.
(Newer open-source packages,
like Boilerpipe and Newspaper, 
exist for text extraction, but we observed they often failed on our web data.)

\begin{table*}[t]
\centering
\begin{tabular}{  l  r  r  r r | r r r r}
\toprule
Model & AUPRC & SE-1 & SE-2 & SE-3 & F1 & SE-1 & SE-2 & SE-3  \\
\midrule
(m1) hard-LR, dep.\ feats. & 0.117 
&
(0.018)
&
(0.005)
&
(0.004)
& 0.229 
&
(0.021)
&
(0.009)
&
(0.008) \\
(m2) hard-LR, n-gram\ feats. & 0.134 
&
(0.020)
&
(0.006)
&
(0.005)
 &0.257 
 &
(0.022)
&
(0.011)
&
(0.009)\\
(m3) hard-LR, all feats. & 0.142
&
(0.021)
&
(0.006)
&
(0.005) 
 & 0.266 
 &
(0.023)
&
(0.010)
&
(0.009)\\
(m4) hard-CNN & 0.130
&
(0.019)
&
(0.006)
&
(0.005)
 & 0.252
 &
(0.022)
&
(0.009)
&
(0.009) \\
\midrule
(m5) soft-CNN (EM) & 0.164
&
(0.023)
&
(0.007)
&
(0.007) 
 & 0.267 
 &
(0.023)
&
(0.009)
&
(0.009)\\
\textbf{(m6) soft-LR (EM)} &  \textbf{0.193}
&
(0.025)
&
(0.008)
&
(0.008) 
 & \textbf{0.316} 
 &
(0.025)
&
(0.011)
&
(0.010) \\
\toprule
Data upper bound (\S\ref{ss:eval}) & 0.57 & -- &-- & -- & 0.73 & -- &-- & --
\\
\toprule
\end{tabular}
\caption{Area under precision-recall curve (AUPRC) and F1 (its maximum value from the PR curve)
for entity prediction on the test set with bootstrap standard errors (SE) sampling from (1) entities (2) documents (3) documents without replacement.
\label{t:results-se}}
\end{table*}

\begin{table}[h]
\centering
\begin{subtable}{\columnwidth}
\centering
\begin{tabular}{l r r r r r}
& m2 & m3 & m4 & m5 & m6 \\
\hline
m1&2.7e-1&1.8e-1&3.1e-1&6.0e-2&6.2e-3 \\
m2&&3.8e-1&4.5e-1&1.7e-1&3.2e-2 \\
m3&&&3.3e-1&2.5e-1&5.8e-2 \\
m4&&&&1.4e-1&2.2e-2 \\
m5&&&&&1.9e-1 \\
\end{tabular}
\caption{Entity resampling}
\vspace{0.2in}
\end{subtable}
\vspace{0.2in}
\begin{subtable}{\columnwidth}
\centering
\begin{tabular}{l r r r r r}
& m2 & m3 & m4 & m5 & m6 \\
\hline
m1&3.5e-2&1.7e-3&5.0e-2&0&0 \\
m2&&1.8e-1&4.1e-1&3.6e-3& 0 \\
m3&&&1.2e-1&3.1e-2& 0 \\
m4&&&&2.1e-3&0 \\
m5&&&&&1.2e-2 \\
\end{tabular}
\caption{Document resampling}
\end{subtable}
\begin{subtable}{\columnwidth}
\centering
\begin{tabular}{l r r r r r}
& m2 & m3 & m4 & m5 & m6 \\
\hline
m1&2.2e-2&8.2-4&9.3e-2&1e-4&0 \\
m2&&1.5e-1&2.6e-1&7.3e-3&0 \\
m3&&&4.6e-2&5.9e-2&0 \\
m4&&&&1.6e-3&0 \\
m5&&&&&2.7e-3 \\
\end{tabular}
\caption{Document resampling with deduplication}
\end{subtable}
\caption{One-sided p-values for for the difference between two models using statistic $T_{ij}$ where $AUPRC_{\text{model}~j} >  AUPRC_{\text{model}~i}$;
    each cell in the table shows $\min(p_{ij}, p_{ji})$.
\label{t:pval}
}
\end{table}



\section{Mention-level preprocessing}
\label{ss:mentdata} 
From the corpus of scraped news documents, to create the mention-level dataset 
 (i.e. the set of sentences containing candidate entities) we :
 \begin{enumerate}
 \setlength\itemsep{0em}
\item Apply the Lynx text-based web browser to extract all a webpage's text.
\item Segment sentences in two steps:
\begin{enumerate}
\item Segment documents to fragments of text (typically, paragraphs)
by splitting on Lynx's representation of HTML paragraph, list markers, and other dividers:
double newlines and the characters -,*, $|$, + and \#.
\item Apply spaCy's sentence segmenter (and NLP pipeline) to these paragraph-like text fragments.
\end{enumerate}
\item De-duplicate sentences as described in detail below.
\item Remove sentences that have fewer than 5 tokens or more than 200.
\item Remove entities (and associated mentions) that
\begin{enumerate}
\item Contain punctuation (except for periods, hyphens and apostrophes).
\item Contain numbers.
\item Are one token in length.
\end{enumerate}
\item Strip any ``'s" occurring at the end of named entity spans. 
\item Strip titles (i.e. Ms., Mr. Sgt., Lt.) occurring in entity spans.
    (HAPNIS sometimes identifies these types of titles; this step basically augments its rules.)
\item Filter to mentions that contain at least one police keyword and at least one fatality keyword.  
\end{enumerate}

\noindent

Additionally, we remove literal duplicate sentences from our mention-level dataset, eliminating all but one duplicated sentence. We select the earliest sentence by download time of its scraped webpage.

\section{\emph{Noisyor} numerical stability}
Under ``hard'' training, many entities at test time have probabilities very close to 1;
in some cases, higher than $1-e^{-1000}$. This happens for entities with a very large number of mentions,
where the naive implementation of \emph{noisyor} as $p = 1-\prod_i (1-p_i)$ has numerical underflow, causing many ties with entities having $p=1$.  
In fact, random tie-breaking for ordering these entity predictions
can give moderate variance to the AUPRC.
(Part of the issue is that
floating point numbers have worse tolerance near 1 than near 0.)

Instead, we rank entity predictions by the log of the complement probability (i.e.\ 1000 for $p=1-e^{-1000}$):
\begin{align*}
\log\left(1- P(y_e=1 \mid x_{\mathcal{M}(e)})\right) \\
= \sum_i \log P(z_i=0 \mid x_i) 
\end{align*}
This is more stable, and while there are a small number of ties,
the standard deviation of AUPRC across random tie breakings
is less than $10^{-10}$.

\section{Manual analysis of results}

Manual analysis is available in Table~\ref{f:top20}.

\section{Bootstrap}

We conduct three different methods of bootstrap resampling,
varying the objects being sampled:
\begin{enumerate}
 \setlength\itemsep{0em}
\item Entities  
\item Documents
\item Documents, with deduplication of mentions.\footnote{To implement, we take the 10,000 samples (with replacement) of documents, and reduce them to the unique set of
    drawn documents. This effectively removes duplicate mentions that occur in method 2 when the same document is drawn more than once in a sample.}
\end{enumerate}

We resample both test-set entities and test-set documents because we are
currently unaware of literature that provides reasoning for one over the other,
and both are arguably relevant in our context.
The bootstrap sampling model assumes a given dataset represents a finite sample
from a theoretically infinite population, and asks
what variability there would be if a finite sample were to be drawn again
from the population.  This has different interpretations for
entity and document resampling.
Resampling entities measures robustness
due to variability in the names that occur in the documents. Resampling
documents measures robustness due to variability in our data source---for
example, if our document scraping procedure was altered, or potentially,
if the news generation process was changed.
Since both entities and documents are not i.i.d., these are both dissatisfying
assumptions.

We also conduct resampling of documents with deduplication of mentions
since, during
development, we found our noisy-or metric was sensitive to duplicate
mentions; this deduplication step effectively includes running our analysis pipeline's
sentence deduplication for each bootstrap sample.

In Fig.~\ref{t:results-se}, we augment the results from Fig.~\ref{t:results}
with standard errors calculated from $B=10,000$ bootstrap samples given the
three methods for sampling described above.  Document resampling tends to give
smaller standard errors than entity resampling, which is to be expected since
there is a larger number of documents than entities.  We analyze our results
using the standard errors and significance tests from method 3.

We examine the statistical significance of difference between models with a one-sided hypothesis test. Our statistic is 
\[ T_{ij} = AUPRC_{\text{model}~j} - AUPRC_{\text{model}~i}.\]
We use hypotheses $H_0: T\leq 0$ and $H_1: T> 0$. As above, we take 10,000 bootstrap samples and find $T^b$ statistic of each sample $b \in \{1..10000\}$.
Then we compute p-values
\[ \text{p-value}_{ij} = \frac{ \text{Count}(T^b_{ij} \leq 0)}{10000} .\]

Finally, since in the observed data, one model is better than the other,
we are interested the null hypothesis that
the apparently-worse
model outperforms the apparently-better model.
Therefore the final p-value comparing systems $i$ and $j$
is actually calculated as $\min(p_{ij}, p_{ji})$,
since the different directions correspond to the fraction of bootstrap samples with
$T_{ij}\leq 0$ versus $T_{ij}>0$;
these values are shown in Fig.~\ref{t:pval}.
(Note $p_{ji}=1-p_{ij}$ in expectation.)
While this seems to follow
standard practice in bootstrap hypothesis
testing in NLP \citep{BergKirkpatrick2012Significance},
we note that \citet{MacKinnon2009Bootstrap}
argues to instead multiply that by two
(i.e., calculate $2 \min(p_{ij}, p_{ji})$)
to conduct a two-sided test that correctly gives $p \sim \text{Unif}(0,1)$
when a null hypothesis of equivalent performance is true.

}

\end{document}